\documentclass[10pt]{article} %
\usepackage[preprint]{tmlr}

\usepackage{amsmath,amsfonts,bm}

\def\eqref#1{equation~\ref{#1}}

\def\1{\bm{1}}

\DeclareMathAlphabet{\mathsfit}{\encodingdefault}{\sfdefault}{m}{sl}
\SetMathAlphabet{\mathsfit}{bold}{\encodingdefault}{\sfdefault}{bx}{n}

\usepackage{hyperref}
\usepackage{url}
\usepackage{xspace}
\usepackage{multirow}
\usepackage{paracol}
\usepackage{graphicx}
\newcommand{\methodName}{ARRN\xspace}
\newcommand{\widthDiagram}{
    1.0\columnwidth
}
\newcommand{\widthDiagramAnnex}{
    1.0\columnwidth
}
\newcommand{\widthPlot}{
    1.0\columnwidth
}

\title{Adaptive Resolution Residual Networks — Generalizing Across Resolutions Easily and Efficiently}

\author{\name Léa Demeule \\
    \email lea.demeule@mila.quebec \\
    \addr Mila - Quebec AI Institute, Université de Montréal 
    \AND
    \name Mahtab Sandhu \\
    \email mahtab.sandhu@mila.quebec \\
    \addr Mila - Quebec AI Institute, Université de Montréal 
    \AND
    \name Glen Berseth \\
    \email glen.berseth@mila.quebec \\
    \addr Mila - Quebec AI Institute, Université de Montréal, and CIFAR 
}

\def\month{MM}  %
\def\year{YYYY} %
\def\openreview{\url{https://openreview.net/forum?id=XXXX}} %

\begin{document}

\maketitle

\begin{abstract}
    The majority of signal data captured in the real world uses numerous sensors with different resolutions. 
    In practice, however, most deep learning architectures are \textit{fixed-resolution}; they consider a single resolution at training time and inference time. This is convenient to implement but fails to fully take advantage of the diverse signal data that exists. 
    In contrast, other deep learning architectures are \textit{adaptive-resolution}; they directly allow various resolutions to be processed at training time and inference time. This benefits robustness and computational efficiency but introduces difficult design constraints that hinder mainstream use.
    In this work, we address the shortcomings of both fixed-resolution and adaptive-resolution methods by introducing \textit{Adaptive Resolution Residual Networks} (\methodName{}s), which inherit the advantages of adaptive-resolution methods and the ease of use of fixed-resolution methods. 
    We construct \methodName{}s from \textit{Laplacian residuals}, which serve as generic adaptive-resolution adapters for fixed-resolution layers, and which allow casting high-resolution \methodName{}s into low-resolution \methodName{}s at inference time by simply omitting high-resolution Laplacian residuals, thus reducing computational cost on low-resolution signals without compromising performance.
    We complement this novel component with \textit{Laplacian dropout}, which regularizes for robustness to a distribution of lower resolutions, and which also regularizes for errors that may be induced by approximate smoothing kernels in Laplacian residuals.
    We provide a solid grounding for the advantageous properties of \methodName{}s through a theoretical analysis based on \textit{neural operators}, and empirically show that \methodName{}s embrace the challenge posed by diverse resolutions with greater flexibility, robustness, and computational efficiency.
\end{abstract}

In general, there is no universal resolution for image data or other signal data; there is, instead, a variety of resolutions that are contingent on the sensors used at the time of capture. 
This poses a challenge to deep learning methods, as adapting and generalizing across diverse resolutions is a complex architecture design problem that is often resolved at the cost of ease of use and compatibility.
On one hand, \textit{fixed-resolution} architectures offer a simple paradigm that is attractive from an engineering standpoint, but that falls short when faced with more diverse resolutions. This form of architecture is forced to cast various resolutions to a single resolution using interpolation, which is undesirable as it inflates computational cost when inference resolution is less than training resolution; it conversely discards useful information when inference resolution is greater than training resolution; it also yields brittle generalization when inference resolution mismatches training resolution.
This approach nonetheless maintains its popularity as it imposes no special constraints on architectural design.
On the other hand, \textit{adaptive-resolution} architectures offer an approach that is generally driven by an analytical study of the link between discrete signals and continuous signals \citep{implicitSine, neuralOperatorFourier, neuralOperator}. 
This approach can guarantee more robust adaptation to various resolution and enables advantageous computational scaling with resolution.
This approach has nonetheless been sparsely used as it has been broadly incompatible with mainstream layer types \citep{neuralOperatorDiscretizationInvariance}.

We propose a class of deep learning architectures that possess both the simplicity of fixed-resolution methods and the computational efficiency and robustness of adaptive-resolution methods. We build these architectures from two simple components: \textit{Laplacian residuals}, which embed standard fixed-resolution layers in architectures that have adaptive-resolution capability, and \textit{Laplacian dropout}, which acts both as a regularizer promoting strong lower resolution performance, and as a regularizer for countering numerical errors that may be induced by approximate implementations of our method.

We situate \methodName{}s relative to prior works and provide background on Laplacian pyramids in \autoref{related} and \autoref{background}. 
We provide an overview of the fundamental notions of signals that allow our work to be formulated and thouroughly define our notation in \autoref{background:signals}. 
We introduce Laplacian residuals and prove that high-resolution \methodName{}s can shed high-resolution Laplacian residuals at inference time to yield low-resolution \methodName{}s that are computationally cheaper yet numerically identical when evaluated on low-resolution signals in \autoref{method:laplacianResidual}. 
We formulate Laplacian dropout in \autoref{method:laplacianDropout} by leveraging the converse idea that randomly omitting Laplacian residuals during training is equivalent to regularizing for robustness using a training distribution that includes lower resolutions; we also note this has a regularizing effect on numerical errors that may be produced by approximate smoothing kernels. 
We perform a set of experiments showing (\autoref{experiments:robustness}) that our method yields stronger robustness at lower resolutions compared to typical fixed-resolution models; (\autoref{experiments:efficiency}) that our method enables significant computational savings through adaptation; (\autoref{experiments:generalization}) that our method is capable of generalizing across layer types in a way that far surpasses prior adaptive-resolution architectures; (\autoref{experiments:adaptationPerfect}) that our theoretical proof for adaptation using ideal smoothing kernels holds empirically; (\autoref{experiments:adaptationDual}) that our theoretical interpretation of the dual regularizing effect of Laplacian dropout using approximate smoothing kernels also holds empirically. 

\section{Related Works}
\label{related}

Here we review related works that allow the formulation of adaptive-resolution architectures. In \autoref{related:adaptiveDomain}, we provide a complimentary discussion of adaptive-resolution methods that are not directly motivated by the interplay between continuous signals and discrete signals.

\paragraph{Adaptive-resolution through neural operators.}
\label{related:adaptiveNeuralOperator}
\citet{neuralOperatorFourier, neuralOperator, neuralOperatorSpectral, neuralOperatorDiscretizationInvariance} are neural architectures that act as \textit{operators} mapping between inputs and outputs that are both \textit{functions}. 
These methods are uniquely defined by their ability to equivalently operate on \textit{continuous functions} or \textit{discrete functions} of any specific resolution.
For every layer that composes a neural operator, there must exists a translation process between a \textit{continuous operator form} and a \textit{discrete operator form} that respects an equivalence constraint which guarantees generalization across different resolutions \citep{neuralOperatorDiscretizationInvariance}. 
For many layers such as standard convolutional layers, transformer layers and adaptive pooling layers, the equivalence constraint is violated, making them incompatible with this technique without substantial alteration \citep{neuralOperatorDiscretizationInvariance}. 
Our method is a form of neural operator, but unlike prior methods, it does not require the layers to conform to this difficult constraint, as Laplacian residuals guarantee their satisfaction independently of the choice of layers.

\paragraph{Adaptive-resolution through implicit neural representations.} 
\label{related:adaptiveImplicitNeuralRepresentation} 
\citet{implicitSignedDistance, implicitOccupancy, implicitSine, implicitRadiance, implicitImageRepresentation, implicitTransformer, implicitTextureEstimator, implicitConv} are neural architectures whose inputs and outputs are also \textit{functions}.
These methods and distinguished by way they implicitly manipulate the functions they operate on; they conceptualize \textit{discrete functions} as samples of underlying \textit{continuous functions} that are approximately reconstructed by \textit{neurally parameterized functions}. 
Some of these methods directly leverage the neurally parameterized functions to allow evaluation at unseen sample location, which has proven highly successful for reconstructing volumetric data, image data and light field data \citep{implicitSignedDistance, implicitOccupancy, implicitSine, implicitRadiance}. 
Some variants of these methods instead derive an alternate representation for the manipulation and generation of functions by separating the parameter space of neurally parameterized functions into a part that is shared across all data points reconstructed during training, which provides a common structure, and a part that is specific to each individual data point reconstructed during training, which acts as a latent representation. 
This form of representation has been successfully used in super-resolution methods for image data \citep{implicitImageRepresentation, implicitTextureEstimator, implicitTransformer}. This form of representation has severe shortcoming when it comes to mapping functions to functions in a resolution-agnostic way, as is required in classification, segmentation or diffusion on image data. Since the symmetries of signals are entangled in latent representations, the incorporation of simple inductive biases on signals becomes problematic, and the majority of mainstream layers tend to be unsuitable \citep{implicitConv}. 
Our method uses the most ubiquitous form of signal representation which is immediately compatible with mainstream layers.
\paragraph{Residual connections}
\label{related:similarResidual}
\citet{residualFrequencyDisentangled} incorporates filtering operations within residuals to separate the frequency content of convolutional networks, although it provides no mechanism for adaptive resolution. \citet{laplacianSuperResolution} uses Laplacian pyramids to solve super-resolution tasks with adaptive \textit{output} resolution, with residuals ordered by \textit{increasing} resolution. 
This is unlike our method, which is well suited to tasks with adaptive \textit{input} resolution, with residuals ordered by \textit{decreasing} resolution. 
\label{related:similarDropout}
\citet{residualStochasticDepth} implements a form of dropout where the layers nested within residual blocks may be bypassed randomly. 
This is somewhat similar to Laplacian dropout, however, this is not equivalent to a form of bandwidth augmentation and does not result in the same improved robustness to various resolutions we show in \autoref{experiments:robustness}.
\section{Background}
\label{background}
In our overview of background notions, we introduce Laplacian pyramids as a stepping stone towards the formulation of Laplacian residuals. In addition, we provide a discussion of signals in \autoref{background:signals} that introduces the notation and fundamental concepts behind this work in a way that is broadly accessible to readers unfamiliar with these ideas, and that should also clarify details valued by more experienced readers.
\subsection{Laplacian pyramids}
\label{background:laplacianPyramid}
In this section, we introduce Laplacian pyramids \citep{laplacianOriginal}, as they have often been used in vision techniques to decompose signals across a range of resolutions, and as they closely relate to Laplacian residuals. 

Laplacian pyramids take some signal $s$, and perform a series of $m$ convolutions with smoothing kernels $\phi_n$ to generate lower bandwidth signals $p^\text{low}_n$ that each are incrementally smoother than the previous one. Laplacian pyramids then generate difference signals $p^\text{diff}_n$ that isolate the part of the signal that was lost at each incremental bandwidth reduction, which intuitively correspond to a certain level of detail of the original signal. The operations that compose a Laplacian pyramid can be captured by a base definition and two simple recursive definitions:
\begin{align}
    \label{equation:laplacianBase}
    p^\text{low}_0 &= s \ast \phi_1 &\in S_1 \\
    \label{equation:laplacianFilterLow}
    p^\text{low}_n &= p^\text{low}_{n - 1} \ast \phi_{n + 1} &\in S_{n + 1} \\
    \label{equation:laplacianFilterBand}
    p^\text{diff}_n &= p^\text{low}_{n - 1} - p^\text{low}_n &\in S_n \setminus S_{n + 1}
\end{align}
Laplacian pyramids are convenient to implement, since each reduction in bandwidth allows a reduction in resolution, as hinted by the notation above, which explicitly includes the bandwidth $S_n$ to highlight the coinciding resolution $|X_n|$ that may be used. 

Laplacian pyramids also allow reconstructing the original signal up to an arbitrary bandwidth using only the last lower bandwidth signal $p^\text{low}_m$ and a variable number of difference signals $p^\text{diff}_n$:
\begin{align}
    \label{equation:laplacianDecompose}
    s \ast \phi_n =  
    p^\text{diff}_{n} +
    p^\text{diff}_{n + 1} +
    \cdots +
    p^\text{diff}_{m - 1} +
    p^\text{diff}_{m} +
    p^\text{low}_{m} \in S_n
\end{align}

\begin{figure*}[h!]
    \centering
    \includegraphics[width=\widthDiagram, page=1]{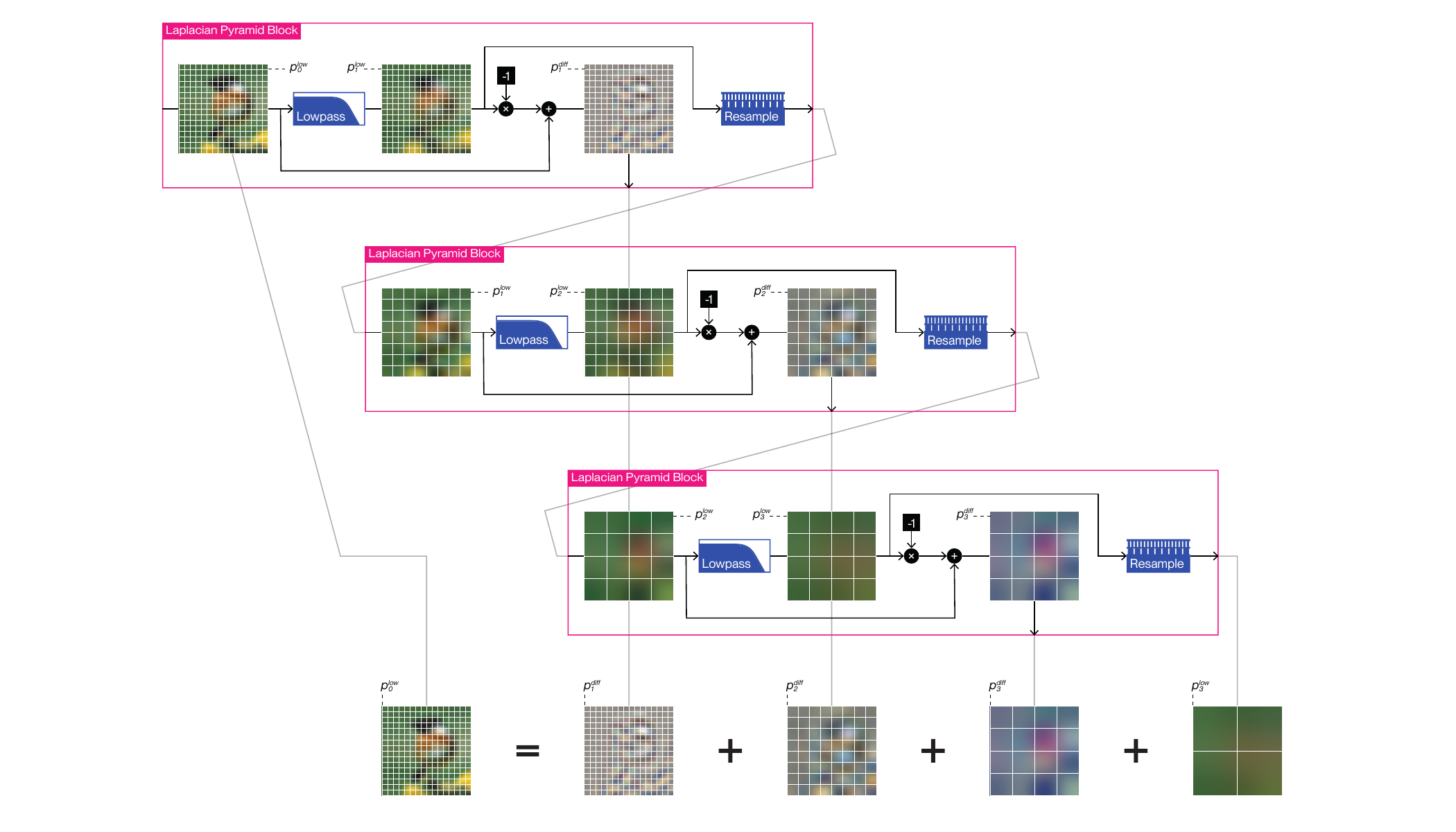}
    \vspace{-2em}
    \caption{
        Visualization of a Laplacian pyramid showing blocks that correspond to recursion iterations. 
        The starting signal $p^\text{low}_{0}$ is shown at the top left. 
        The blocks are chained one after the other, from top-to-bottom, and produce the difference signals $p^\text{diff}_{1}$, $p^\text{diff}_{2}$, $p^\text{diff}_{3}$.
        The blocks each start with the signal $p^\text{low}_{n - 1}$, produce a lower bandwidth signal $p^\text{low}_{n}$ moving right, and finally produce a difference signal $p^\text{diff}_{n}$ moving right again.
        Together, $p^\text{diff}_{1}$, $p^\text{diff}_{2}$, $p^\text{diff}_{3}$ and $p^\text{low}_{3}$ sum to the original signal $p^\text{low}_{0}$; the Laplacian pyramid produces a form of linear decomposition.
    }
    \label{figure:laplacianPyramid}
\end{figure*}

In \autoref{figure:laplacianPyramid}, we summarize the recursive formulation of Laplacian pyramids in a three-block pyramid; this is intended to allow easy comparison with the Laplacian residuals we illustrate in \autoref{figure:laplacianResidual}.

\newpage

Laplacian pyramids are typically constructed using Gaussian smoothing kernels, which violates \autoref{equation:discretizationCorrectness}, and introduces errors in the sampling process as a consequence. We require perfect Whittaker-Shannon smoothing kernels for the exact computation reduction property we show in Laplacian pyramids just below and in Laplacian residuals later in \autoref{method:laplacianResidual}, as these smoothing kernels are uniquely able to satisfy \autoref{equation:discretizationInvariance}. However, we show in \autoref{experiments:adaptationDual} that Laplacian dropout enables \methodName{}s to learn to correct these approximation errors.

\columnratio{0.5}
\begin{paracol}{2}
    \begin{nthcolumn}{0}
        Laplacian pyramids have the desirable ability to adapt to low resolutions by skipping computations, as we may intuit from \autoref{equation:laplacianDecompose} and \autoref{figure:laplacianPyramid}. We formally show this by considering a signal in discrete form $s[\ \cdot \ ]_u$ with a low resolution $|X_u|$ and a corresponding low bandwidth constraint $S_u$. We are interested in what happens up to the level $n$ of the Laplacian pyramid, where $n$ is the highest level that sits just at or above the resolution $|X_n| \geq |X_u|$ and the bandwidth constraint $S_n \supseteq S_u$ of the original signal. We observe that all of the smoothing filters $\phi_1, \cdots, \phi_n$ associated with prior levels of the Laplacian pyramids leave the original signal $s$ unchanged since $s \in S_u \subseteq S_n$ and since $S_n \subset \cdots \subset S_1$, which induces a trail of zero terms in the expansion of the recursive terms \autoref{equation:laplacianFilterLow} and \autoref{equation:laplacianFilterBand}, as shown in \autoref{equation:laplacianAdaptation} on the right.
    \end{nthcolumn}
    \begin{nthcolumn}{1}
        \begin{align}
            \implies p^\text{low}_{0} &= s \ast \phi_{1} \\
            &= s \\
            \implies p^\text{low}_{1} &= p^\text{low}_{0} \ast \phi_{2} \\
            &= s \\
            \implies p^\text{diff}_{1} &= p^\text{low}_{0} - p^\text{low}_{1} \\
            &= 0 \\
            & \vdots \notag \\
            \implies p^\text{low}_{n - 1} &= p^\text{low}_{n - 2} \ast \phi_{n} \\
            &= s \\
            \implies p^\text{diff}_{n - 1} &= p^\text{low}_{n - 2} - p^\text{low}_{n - 1} \\
            \label{equation:laplacianAdaptation}
            &= 0
        \end{align}
    \end{nthcolumn}
\end{paracol}
This allows directly setting $p^\text{low}_{n - 1} = s$ and carrying out computation only to recover difference terms starting at $p^\text{diff}_n$, skipping all difference terms $p^\text{diff}_{1}, \ \dots \ , p^\text{diff}_{n - 1}$. We later design Laplacian residuals to reproduce exactly this desirable behaviour such that we can adapt deep learning architectures to lower resolutions by skipping computations.
\section{Method}
\label{method}
In this section, we build towards \textit{Laplacian residuals} (\autoref{method:laplacianResidual}), which are designed to allow the construction of adaptive-resolution architectures from standard fixed-resolution layers, and \textit{Laplacian dropout} (\autoref{method:laplacianDropout}), which both serves as a regularizer for robustness at lower resolution, and a regularizer for error-correction when imperfect smoothing kernels are used.
\subsection{Laplacian residuals for adaptive-resolution deep learning}
\label{method:laplacianResidual}
Laplacian residuals are alike to Laplacian pyramids in the way they separate a signal into a sum of progressively lower bandwidth signals, and in the way they are able to operate at lower resolution by simply skipping computations. However, Laplacian residuals crucially differ in their ability to incorporate neural architectural blocks that enable deep learning.

\begin{figure*}[h!]
    \centering
    \includegraphics[width=\widthDiagram, page=2]{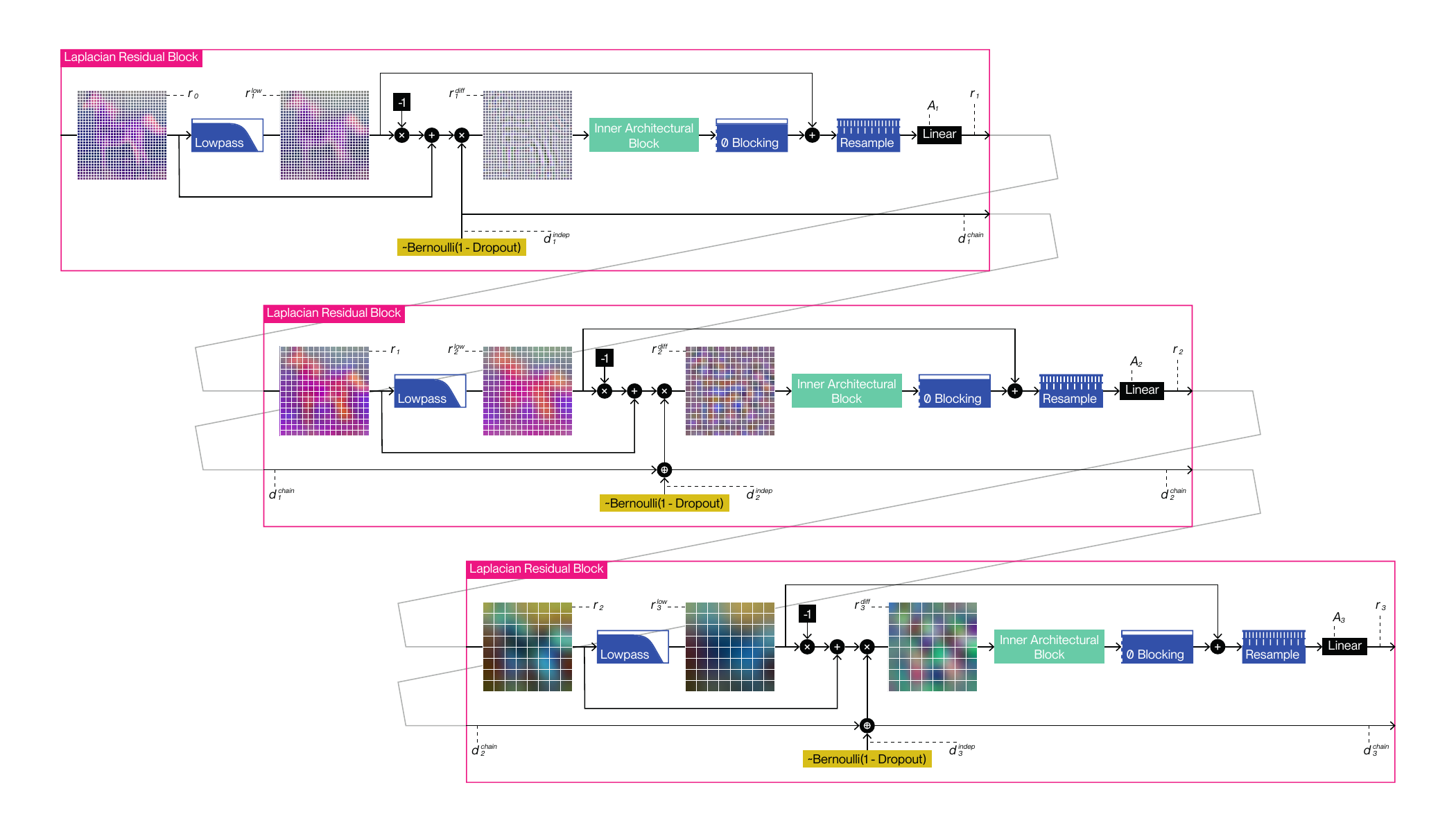}
    \vspace{-2em}
    \caption{
        Visualization of an \methodName{} showing blocks that correspond to Laplacian residuals, displaying a principal component analysis of feature maps.
        The starting signal $r_{0}$ is shown at the top left. 
        The blocks are chained one after the other, from top-to-bottom.
        The blocks each start with the previous residual signal $r_{n - 1}$, and, moving left-to-right, produce a lower bandwidth signal $r^\text{low}_n$ and difference signal $r^\text{diff}_n$ as in Laplacian pyramids, gate the difference signal against the chained Laplacian dropout variable $d^\text{chain}_n$, apply the neural architectural block $b_{n}$, apply the zero-blocking filter $\psi$, sum the lower bandwidth signal $r^\text{low}_n$, apply the smoothing kernel $\phi_{n + 1}$ and resample, then finally project through $\mathbf{A}_n$ to produce the residual signal $r_{n}$.
    }
    \label{figure:laplacianResidual}
\end{figure*}

Laplacian residuals are formulated as adaptive-resolution layers \textit{(operators on continuous signals)} $r_n : (\mathbf{X} \to \mathbb{R}^{f_n}) \in S_n \to (\mathbf{X} \to \mathbb{R}^{f_{n + 1}}) \in S_{n + 1}$ that incorporate neural architectural blocks $b_n : (\mathbf{X}_n \to \mathbb{R}^{f_n}) \to (\mathbf{X}_n \to \mathbb{R}^{f_n})$ that are fixed-resolution layers \textit{(operators on discrete signals)}. This formulation enables wide compatibility with mainstream fixed-resolution layers that is unseen in prior adaptive-resolution methods. This compatibility is only conditional on the neural architectural block $b_n$ producing a constant signal when its input is zero \autoref{equation:residualInnerConstraint}, which is trivially guaranteed by linear layers, activation layers, convolutional layers, batch normalization layers, some transformer layers, and any composition of layers that individually meet this condition:
\begin{align}
    \label{equation:residualInnerConstraint}
    b_n\{0\} = a \ \text{where} \ a \in \mathbb{R}^{f_n}
\end{align}
The base case and recursive cases seen in Laplacian residuals are nearly identical to those of Laplacian pyramids, aside from including a linear projection $\mathbf{A}_0 : \mathbb{R}^{f_1 \times f_0}$ to raise the feature dimensionality from $f_0 \in \mathbb{N}$ to $f_1 \in \mathbb{N}$ before the first neural architectural block:
\begin{align}
    \label{equation:residualBase}
    r_0 &= \mathbf{A}_0 s \ast \phi_1 &\in S_1 \\
    \label{equation:residualFilterLow}
    r^\text{low}_n &= r_{n - 1} \ast \phi_{n + 1} &\in S_{n + 1} \\
    \label{equation:residualFilterBand}
    r^\text{diff}_n &= r_{n - 1} - r^\text{low}_n &\in S_n \setminus S_{n + 1}
\end{align}
The neural architectural block $b_n$ receives the difference signal $r^\text{diff}_n$ as its input, and then sums its output with the lower bandwidth signal $r^\text{low}_n$ before passing it to the next Laplacian residual. The signals are combined with some additional processing, which we define below and motivate more concretely next:
\begin{align}
    \label{equation:residualInner}
    r_n = \mathbf{A}_n (
        b_n\{
            r^\text{diff}_n
        \} \ast \psi \ast \phi_{n + 1}
        + r^\text{low}_n
    ) \in S_{n + 1}
\end{align}
In \autoref{figure:laplacianResidual}, we summarize the recursive formulation of Laplacian residuals into a diagram that illustrates a three-block \methodName{}, which allows easy comparison with the Laplacian pyramid shown in \autoref{figure:laplacianPyramid}. 

We include the kernel $\psi$ in \autoref{equation:residualInner} to allow Laplacian residuals to replicate the same computation skipping behaviour seen in Laplacian pyramids. We achieve this by setting the kernel $\psi$ to a constant rejection kernel; the convolution against $\psi$ effectively subtracts the mean and ensures the neural architectural block $b_n$ contributes zero to the residual signal $r_n$ if the difference signal $r^\text{diff}_n$ is zero, as shown in \autoref{equation:residualInnerZeroBlock}. We obtain this result thanks to the constraint set on the neural architectural block $b_n$ in \autoref{equation:residualInnerConstraint}:
\begin{align}
    \label{equation:residualInnerZeroBlock}
    b_n\{0\} \ast \psi = 0
\end{align}
We include a smoothing kernel $\phi_{n + 1}$ in \autoref{equation:residualInner} to ensure the output bandwidth of $r_n$ coincides with the input bandwidth of $r_{n + 1}$, which is necessary for Laplacian residuals to follow the same general structure as Laplacian pyramids.

We also incorporate a projection matrix $\mathbf{A}_n : \mathbb{R}^{f_{n + 1} \times f_n}$ in \autoref{equation:residualInner} to allow raising the feature dimensionality from $f_n \in \mathbb{N}$ to $f_{n + 1} \in \mathbb{N}$ at the end of each Laplacian residual, so that more capacity can be allocated to later Laplacian residuals.

We note that the neural architectural block $b_n\{\ \cdot \ \}$ in \autoref{equation:residualInner} is written in shorthand, and stands for the more terse expression $\mathfrak{I}_n\{b_n\{\mathfrak{S}_n\{\ \cdot \ \}\}\}$. This is a formal trick that enables our analysis by casting the neural architectural block from an \textit{operator on discrete signals} $(\mathbf{X}_n \to \mathbb{R}^{f_n}) \to (\mathbf{X}_n \to \mathbb{R}^{f_n})$ to an \textit{operator on continuous signals} $(\mathbf{X} \to \mathbb{R}^{f_n})  \in S_n \to (\mathbf{X} \to \mathbb{R}^{f_n})  \in S_n$ that is equivalent in the sense of \autoref{equation:discretizationInvariance}. 
This does not directly reflect the implementation of the method, as all linear operators are analytically composed then cast to their discrete form to maximize computational efficiency.

We add that we can alter the formulation of \autoref{equation:residualInner} to let the neural architectural block perform a parameterized downsampling operation $(\mathbf{X}_n \to \mathbb{R}^{f_n}) \to (\mathbf{X}_{n + 1} \to \mathbb{R}^{f_n})$ by changing the interpolation operator discussed above from $\mathfrak{I}_{n} : (\mathbf{X}_{n} \to \mathbb{R}^{f_n}) \to (\mathbf{X} \to \mathbb{R}^{f_n})  \in S_{n}$ to $\mathfrak{I}_{n + 1} : (\mathbf{X}_{n + 1} \to \mathbb{R}^{f_n}) \to (\mathbf{X} \to \mathbb{R}^{f_n})  \in S_{n + 1}$ and by dropping the smoothing kernel $\phi_{n + 1}$ from \autoref{equation:residualInner}. 

\paragraph{Adaptation to lower resolution signals with perfect smoothing kernels.} We guarantee that architectures built from a series of Laplacian residuals can adapt to lower resolution signals by simply skipping the computation of higher resolution Laplacian residuals \textit{without causing any numerical perturbation} when using \textit{perfect} smoothing kernels. This guarantee provides strong theoretical backing to the validity of our method, and is also supported by empirical evidence in \autoref{experiments:adaptationPerfect}.

We show this property following the same argument leveraged to derive \autoref{equation:laplacianAdaptation} in the case of Laplacian pyramids. We take a signal in discrete form $s[\ \cdot \ ]_u$ with a low resolution $|X_u|$ and a corresponding low bandwidth constraint $S_u$. We consider Laplacian residuals up to the level $n$ of the architecture, where $n$ is the highest level that sits just at or above the resolution $|X_n| \geq |X_u|$ and the bandwidth constraint $S_n \supseteq S_u$ of the original signal. We observe that all of the smoothing filters $\phi_1, \cdots, \phi_n$ associated with prior levels of the architecture leave the original signal $s$ unchanged since $s \in S_u \subseteq S_n$ and since $S_n \subset \cdots \subset S_1$, which induces a trail of zero terms in the expansion of the recursive terms \autoref{equation:residualFilterLow}, \autoref{equation:residualFilterBand}, and \autoref{equation:residualInner}, as shown in \autoref{equation:residualAdaptation}:
\begin{align}
    \implies r^\text{low}_{0} &= \mathbf{A}_0 s \ast \phi_{1} \\
    &= \mathbf{A}_0 s \\
    \implies r^\text{low}_{1} &= r_{0} \ast \phi_{2} \\
    &= \mathbf{A}_0 s \\
    \implies r^\text{diff}_{1} &= r_{0} - r^\text{low}_{1} \\
    &= 0 \\
    \implies r_{1} &= \mathbf{A}_{1} (
        b_{1}\{
            r^\text{diff}_{1}
        \} \ast \psi \ast \phi_{2}
        + r^\text{low}_{1}
    ) \\
    &= \mathbf{A}_1 \mathbf{A}_0 s \\
    \vdots & \notag \\
    \implies r^\text{low}_{n - 1} &= r_{n - 2} \ast \phi_{n} \\
    &= \mathbf{A}_{n - 2} \cdots \mathbf{A}_0 s \\
    \implies r^\text{diff}_{n - 1} &= r_{n - 2} - r^\text{low}_{n - 1} \\
    &= 0 \\
    \implies r_{n - 1} &= \mathbf{A}_{n - 1} (
        b_{n - 1}\{
            r^\text{diff}_{n - 1}
        \} \ast \psi \ast \phi_{n}
        + r^\text{low}_{n - 1}
    ) \\
    \label{equation:residualAdaptation}
    &= \mathbf{A}_{n - 1} \cdots \mathbf{A}_0 s
\end{align}
This can be leveraged to evaluate a chain of Laplacian residuals at a lower resolution by simply discarding higher-resolution Laplacian residuals and only considering the chain of linear projections $\mathbf{A}_{n - 1} \cdots \mathbf{A}_0$ that is carried over. This provides adaptive-resolution capability to fixed-resolution layers without difficult design constraints, and improves computation efficiency without compromising numerical accuracy in any way.

We can use this result to state an equivalence between evaluation using all Laplacian residuals (\autoref{equation:residualEquivalenceAll}) and evaluation using the strictly necessary Laplacian residuals (\autoref{equation:residualEquivalenceAdapted}):
\begin{align}
    \label{equation:residualEquivalenceAll}
    & \ \mathfrak{S}_m\{
        r_m\{\ \cdots \ r_0\{
            \mathfrak{S}_0\{
                \mathfrak{I}_{u}\{
                    s[\ \cdot \ ]_u
                \}
            \}
        \} \ \cdots \ \}
    \} \\
    \label{equation:residualEquivalenceAdapted}
    =& \ \mathfrak{S}_m\{
        r_m\{\ \cdots \ r_n\{
            \mathbf{A}_{n - 1} \cdots \mathbf{A}_0
            \mathfrak{S}_n\{
                \mathfrak{I}_{u}\{
                    s[\ \cdot \ ]_u
                \}
            \}
        \} \ \cdots \ \}
    \}
\end{align}

\paragraph{Adaptation to lower resolution signals with approximate smoothing kernels.}
We show that using \textit{approximate} smoothing kernels \textit{causes some numerical perturbation} when skipping the computation of higher resolution Laplacian residuals. This observation motivates the use of Laplacian dropout, a training augmentation we introduce in \autoref{method:laplacianDropout} that addresses this limitation while also improving robustness.

When using imperfect smoothing kernels $\rho_n \approx \phi_n$ , the guarantee we provide does not hold exactly. We consider the case case where $\phi_n$ would leave a signal $s$ unchanged, and note that $\rho_n$ would disturb the signal $s$ by a small error signal $\epsilon_n$:
\begin{align}
    \label{equation:residualKernelError}
    s \ast \phi_n = s \implies s \ast \rho_n = s + \underbrace{s \ast (\rho_n - \phi_n)}_{\epsilon_n}
\end{align}
We highlight that the discrepancy above would induce a small error term $\epsilon_n$ in every intermediate zero term that leads to \autoref{equation:residualAdaptation}, and therefore discarding the Laplacian residuals (\autoref{equation:residualEquivalenceAdapted}) would not be exactly equivalent to retaining all Laplacian residuals (\autoref{equation:residualEquivalenceAll}). We note that this is not simply constrained to a linear effect, as $\epsilon_1$ will for instance affect $b_1$, which has nonlinear behavior.
\subsection{Laplacian dropout for effective generalization}
\label{method:laplacianDropout}
In this section, we introduce Laplacian dropout, a training augmentation that is specially tailored to improve the performance of our method by taking advantage of the structure of Laplacian residuals, and that comes at effectively no computational cost.

We formulate Laplacian dropout by following the intuition that Laplacian residuals can be randomly disabled during training to improve generalization. We only allow disabling consecutive Laplacian residuals using the logical or operator to ensure that Laplacian dropout does not cut intermediate information flow:
\begin{align}
    \label{equation:residualDropoutIndep}
    d^\text{indep}_n &\sim B(1 - p_n) \\
    \label{equation:residualDropoutChain}
    d^\text{chain}_n &= d^\text{indep}_n \oplus d^\text{chain}_{n - 1} \\
    \label{equation:residualFilterBandDropout}
    r^\text{diff}_n &= d^\text{chain}_n (r_{n - 1} - r^\text{low}_n)
\end{align}
Next, we provide a theoretical interpretation that identifies two distinct purposes that Laplacian dropout fulfills in our method. We see this dual utility as a highly desirable feature of Laplacian dropout.

\paragraph{Regularization of robustness at lower resolution.} Since Laplacian dropout truncates Laplacian residuals in the same way they are truncated when adapted to lower resolutions, Laplacian dropout is identical to randomly lowering resolution when using perfect smoothing kernels. This acts as a training augmentation that promotes robustness over a \textit{distribution} of lower resolutions. We perform a set of classification tasks in \autoref{experiments:robustness} that show this regularizing effect sometimes \textit{doubling} accuracy over certain lower resolutions without adversely affecting accuracy at the highest resolution.
\paragraph{Regularization of errors introduced by approximate smoothing kernels.} Since Laplacian dropout truncates Laplacian residuals in the same way they are truncated when adapted to lower resolutions, Laplacian dropout exactly replicates numerical errors produced by approximate smoothing kernels in \autoref{equation:residualKernelError}. This allows learning a form of error compensation that offsets the effect of approximate smoothing kernels. We demonstrate this allows the use of very coarsely approximated smoothing kernels in \autoref{experiments:adaptationDual} that otherwise impart a significant performance penalty on our method.

\section{Experiments}
\label{experiments}
We present a set of experiments that demonstrate our method's robustness across resolutions, its computational efficiency, and its ease of use. We show (\autoref{experiments:robustness}) that our method is highly robust across diverse resolutions; (\autoref{experiments:efficiency}) that adaptation provides our method with a significant computational advantage; (\autoref{experiments:generalization}) that our method can generalize across layer types in a way that exceeds the capabilities of prior adaptive-resolution architectures; (\autoref{experiments:adaptationPerfect}) that our theoretical guarantee for adaptation to lower resolutions with perfect smoothing kernels holds empirically; and (\autoref{experiments:adaptationDual}) that our theoretical interpretation of the dual regularization effect of Laplacian dropout coincides with the behaviour we observe empirically. 

\paragraph{Experiment design.}
We compare models in terms of their robustness across resolutions, their computational scaling relative to resolution, and their ease of construction. We follow a typical use case for our method, where we train each model at a single resolution and then evaluate over a range of resolutions. We consider the fluctuation of accuracy and inference time over resolution as the metrics of interest for our discussion. We perform a set of classification tasks that require models to effectively leverage the information of low-resolution to medium-resolution images; \textbf{CIFAR10} ($32 \times 32$) \citep{datasetCIFAR}, \textbf{CIFAR100} ($32 \times 32$) \citep{datasetCIFAR}, \textbf{TinyImageNet} ($64 \times 64$) \citep{datasetTinyImageNet} and \textbf{STL10} ($96 \times 96$) \citep{datasetSTL10}. 

\newpage
\paragraph{Model design and selection.}
We apply our method by constructing adaptive-resolution models that rely on commonly used fixed-resolution layers. For most of our experiments (\autoref{experiments:robustness}, \autoref{experiments:efficiency}, \autoref{experiments:adaptationPerfect} and \autoref{experiments:adaptationDual}), we take inspiration from MobileNetV2 \citep{convMobileNetV2} and EfficientNetV2 \citep{convEfficientNetV2} to construct \methodName{}s that are documented in \autoref{experiments:details}. For the experiment that investigates generalization across layer types (\autoref{experiments:generalization}), we instead construct \methodName{}s by transplanting layers that are found across a range of mainstream fixed-resolution architectures: \textbf{ResNet18}, \textbf{ResNet50}, \textbf{ResNet101} (11.1M-42.5M) \cite{residualOriginal}, \textbf{WideResNet50V2}, \textbf{WideResNet101V2} (66.8M-124M) \citep{convWideResNet}, \textbf{MobileNetV3Small}, \textbf{MobileNetV3Large}(1.52M-4.21M) \cite{convMobileNetV3}. We splice the sequence of layers that composes each fixed-resolution architecture at points where resolution changes occur and nest each resulting subsequence of layers in a Laplacian residual with matching resolution. We discard the first two Laplacian residuals for MobileNetV3, as the resolution of the tailing Laplacian residuals otherwise becomes very small. For our choice of baseline methods, we consider mainstream fixed-resolution architectures to show they compromise robustness across diverse resolutions, yet they have no substantial advantage in ease of implementation or compatibility with existing layers relative to our method, as demonstrated by our experiment on generalization across layer types. We include all the fixed-resolution architectures above in this comparison, along with \textbf{EfficientNetV2S}, \textbf{EfficientNetV2M}, \textbf{EfficientNetV2L} (20.2M-117.2M) \citep{convEfficientNetV2}. For the experiments that validate our theoretical analysis (\autoref{experiments:adaptationPerfect} and \autoref{experiments:adaptationDual}), we perform an ablation study over the quality of the smoothing filter, the use of Laplacian dropout at training time, and the use of adaptation at inference time.

\paragraph{Model training and evaluation.} All models are trained for 100 epochs at the full dataset resolution with identical hyperparameters that are described in \autoref{experiments:details}. All models are then evaluated at the full dataset resolution and at a range of lower resolutions that are synthetically generated. In the case of fixed-resolution models, the lower resolution signals are interpolated to the resolution supported by the models. In the case of our models, the lower resolution signals are directly processed if they coincide with the resolution of a Laplacian residual; if they fall in between the resolution of two successive Laplacian residuals, the higher resolution is generally interpolated to, as required by our theoretical proof (\autoref{equation:residualAdaptation}). However, the lower resolution is sometimes interpolated to instead as it can result in more robust behavior that is computationally cheaper. This is the case with TinyImageNet and STL10 in \autoref{experiments:robustness} and \autoref{experiments:efficiency}, and with TinyImageNet in \autoref{experiments:generalization}. This effect likely results from more consistent statistical properties encountered when only evaluating Laplacian residuals that have full access to the part of the signal they usually address.

\subsection{Robustness and the effectiveness of Laplacian dropout}
\label{experiments:robustness}

\begin{figure*}[h!]
    \centering
    \includegraphics[width=\widthPlot, page=1]{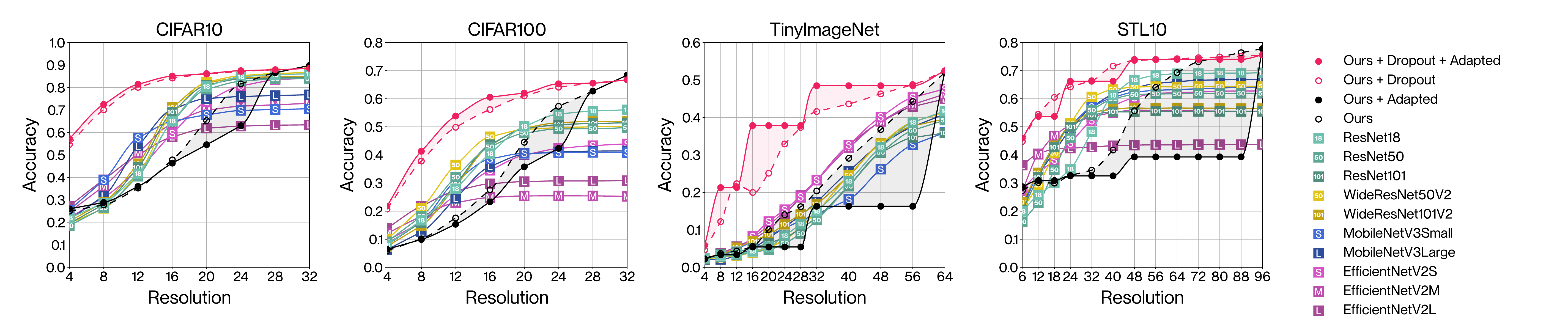}
    \vspace{-2em}
    \caption{
        Accuracy of all methods at various resolutions, where each model is trained at the full dataset resolution and tested at a range of lower resolutions. Our method (red full line) displays the best accuracy at the highest resolution and robustly maintains its accuracy at lower resolutions.
    }
    \label{figure:robustness}
    \vspace{-2em}
\end{figure*}

We demonstrate that our method allows for greater low-resolution robustness than mainstream methods without compromise in high-resolution performance. \autoref{figure:robustness} shows four ablations of our method corresponding to permutations of two sets: either with Laplacian dropout (red lines) or without Laplacian dropout (black lines); and either with adaptation (solid lines) or without adaptation (dashed lines). We see that with Laplacian dropout and with adaptation (full red lines), our method outperforms every baseline method across every resolution and every dataset. We also find that, in contrast, without Laplacian dropout (black lines), our method shows much weaker generalization across resolutions, clearly demonstrating Laplacian dropout's effectiveness as a regularizer for robustness to diverse resolutions.

\subsection{Computational efficiency}
\label{experiments:efficiency}

\begin{figure*}[h!]
    \centering
    \includegraphics[width=\widthPlot, page=2]{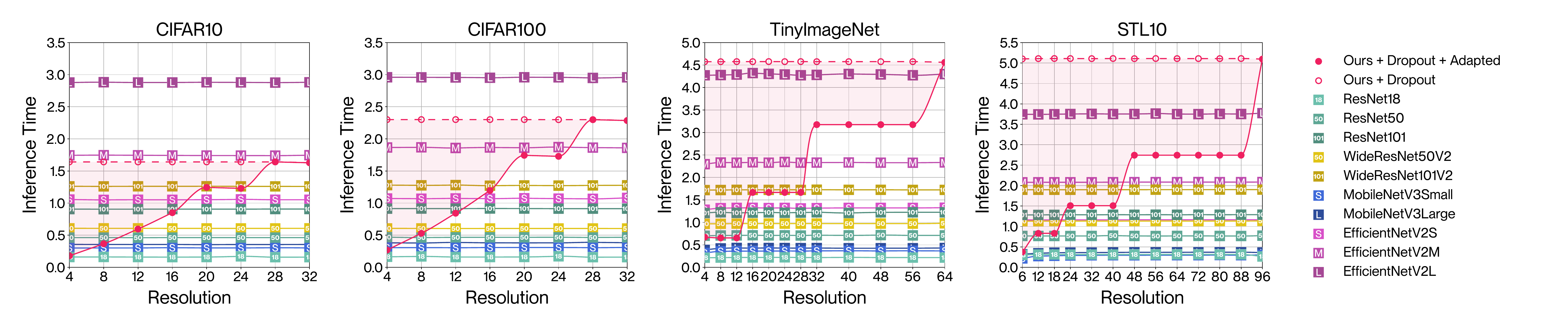}
    \vspace{-2em}
    \caption{
        Inference time of all methods at various resolutions, where the inference time for the entire dataset is considered. Our method (red full line) can adapt to lower resolutions by skipping Laplacian residuals, which results in significant computational savings (highlighted by the shaded area) compared to using all Laplacian residuals (red dashed line). Our method also displays a reasonable inference time relative to typical convolutional neural networks despite not having a highly optimized implementation.
    }
    \label{figure:timeInference}
\end{figure*}

We confirm the computational savings granted by adaptation by performing time measurements on the previous experiment. We use CUDA event timers and CUDA synchronization barriers around the forward pass of the network to eliminate other sources of overhead, such as data loading, and sum these time increments over all batches of the full dataset. 
We repeat this process $10$ times and pick the median to reduce the effect of outliers. \autoref{figure:timeInference} shows the inference time of \methodName{}s with adaptation (full red lines) and without adaptation (dashed red lines). Our method significantly reduces its computational cost (highlighted by the shaded area) by requiring the evaluation of a lower number of Laplacian residuals at lower resolutions. Our method also has a reasonable inference time relative to well-engineered standard methods overall.

\subsection{Generalization across layer types}
\label{experiments:generalization}
\begin{figure*}[htb]
    \centering
    \includegraphics[width=\widthPlot, page=3]{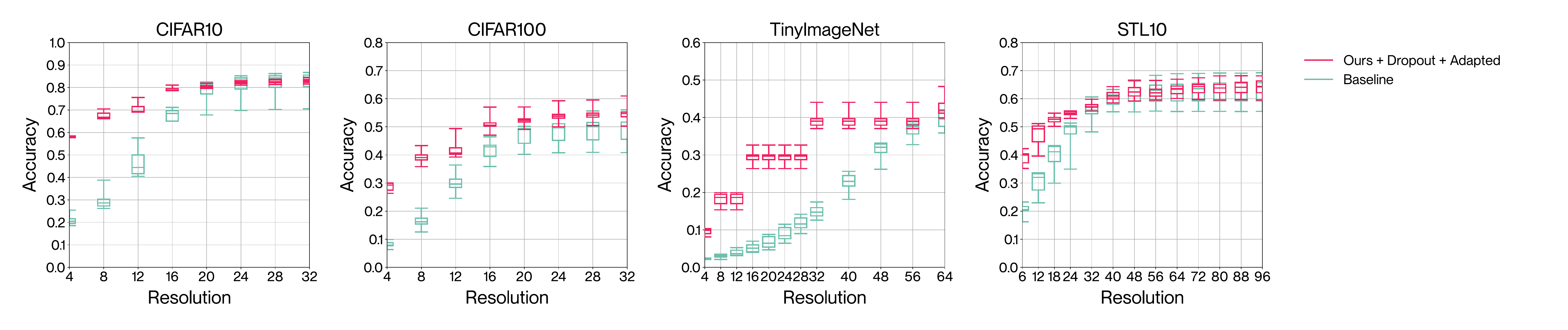}
    \vspace{-2em}
    \caption{
        Accuracy of two groups of methods at various resolutions, where 7 adaptive-resolution architectures (in red) are constructed by taking 7 fixed-resolution architectures (in green) and wrapping their layers in Laplacian residuals. Our method yields architectures that have much stronger low-resolution performance, and similar or better high-resolution performance, which demonstrates ease of use and wide compatibility with mainstream layers.
    }
    \label{figure:generalization}
\end{figure*}

We demonstrate the ease of use of our method and its capability to incorporate various layer types by constructing adaptive-resolution architectures from a range of mainstream fixed-resolution architectures (\textbf{ResNet18}, \textbf{ResNet50}, \textbf{ResNet101}, \textbf{WideResNet50V2}, \textbf{WideResNet101V2}, \textbf{MobileNetV3Small} and \textbf{MobileNetV3Large}). \autoref{figure:generalization} compares the accuracy of architectures in adaptive-resolution form (red box plots) and in fixed-resolution form (green box plots). The distribution of accuracies of the seven underlying architectures in the adaptive-resolution group and fixed-resolution group is visually conveyed in a decluttered format by drawing a small box plot at every resolution. Our method consistently delivers better low-resolution performance, and similar or better high-resolution performance. Our method achieves this while generalizing beyond the abilities of prior adaptive-resolution architectures, as it is able to incorporate the fixed-resolution layers used in this experiment.

\subsection{Adaptation with perfect smoothing kernels}
\label{experiments:adaptationPerfect}

We perform an ablation study to verify our theoretical guarantee for numerically identical adaptation. 
\autoref{figure:ablationGraph} displays a set of experiments that use \textit{perfect quality} Whittaker-Shannon smoothing kernels (in the upper row of graphs in green) implemented through the Fast Fourier Transform \citep{fourierFast}. We showcase the usual set of ablations within this group of experiments; with Laplacian dropout (bright green lines) or without Laplacian dropout (dark green lines); with adaptation (full lines) or without adaptation (dashed lines). Our method evaluates practically identically whether unnecessary Laplacian residuals are discarded (with adaptation, full lines, \autoref{equation:residualEquivalenceAdapted}), or whether all Laplacian residuals are preserved (without adaptation, dashed lines, \autoref{equation:residualEquivalenceAll}), with imperceptible discrepancies that are exactly zero, or that are small enough to be attributed to the numerical limitations of floating point computation. Our method is able to skip computations without numerical compromises, as predicted by our theoretical guarantee. 

\subsection{Adaptation with approximate smoothing kernels and the dual effect of Laplacian dropout}
\label{experiments:adaptationDual}

\begin{figure*}[h!]
    \centering
    \includegraphics[width=\widthPlot, page=4]{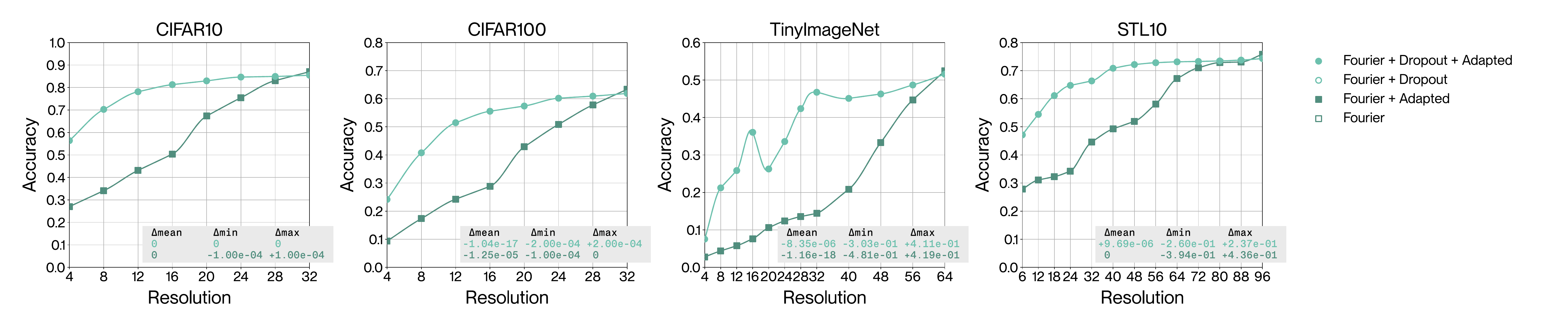}
    \includegraphics[width=\widthPlot, page=5]{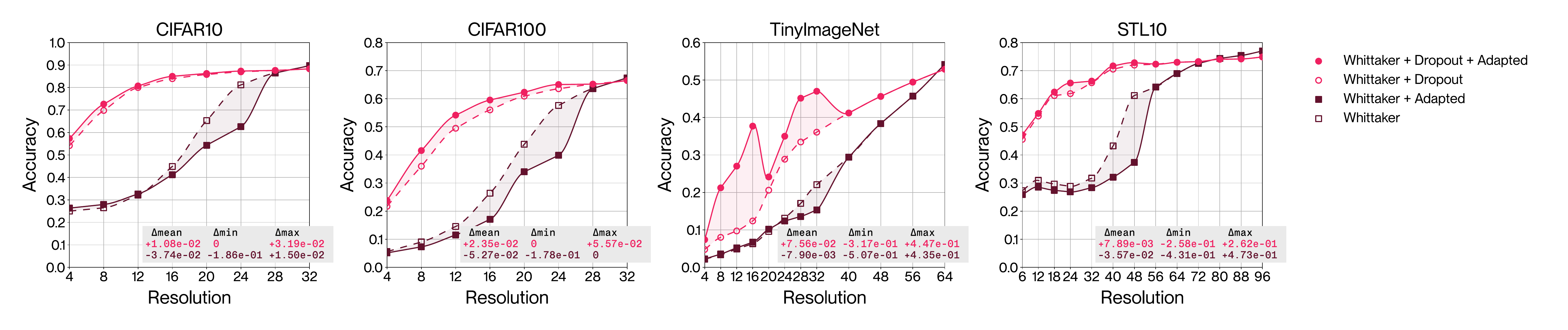}
    \includegraphics[width=\widthPlot, page=6]{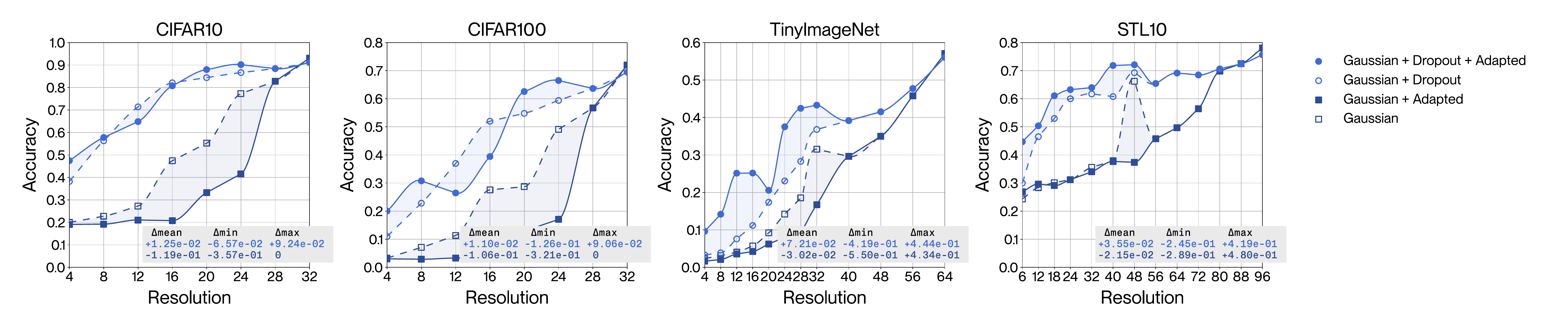}
    \vspace{-2em}
    \caption{
         Accuracy of {\methodName}s at various resolutions, where each \methodName is identical to the architecture used in the main experiments aside from the quality of smoothing kernel. The smoothing kernels each correspond to a different row of graphs and a different hue. The shaded area spanning pairs of accuracy curves highlights the difference between the accuracy of the same model evaluated with adaptation and without adaptation. The tables display a statistical breakdown of this discrepancy for each model.
    }
    \label{figure:ablationGraph}
\end{figure*}

\begin{figure*}[h!]
    \centering
    \includegraphics[width=\textwidth]{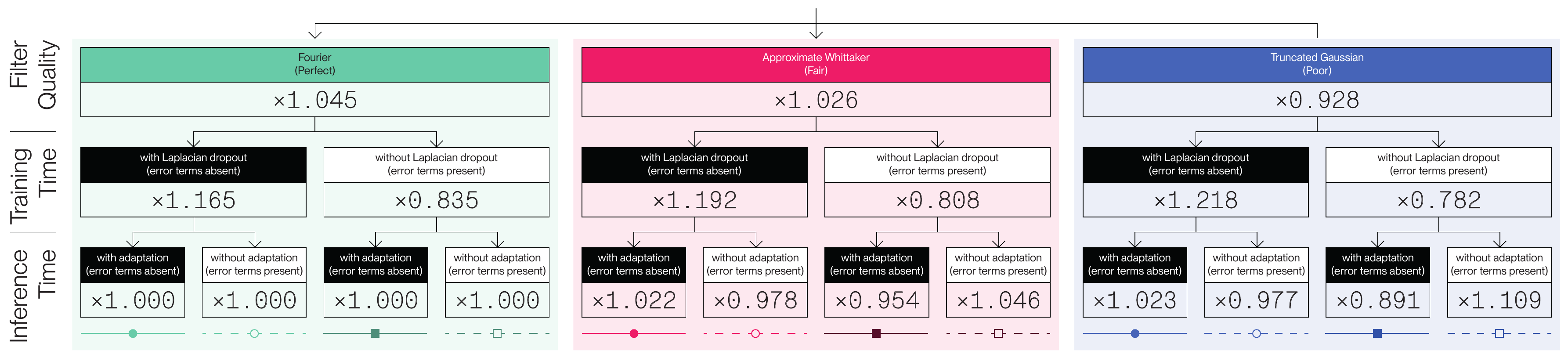}
    \vspace{-1em}
    \caption{
        Multiplicative factor analysis showing an ablation over key components of \methodName{}s. The performance metric considered for analysis is the average accuracy over all resolutions and all datasets. The three levels of the decision tree show which filter is used, whether Laplacian dropout is used at training time, and whether adaptation is used at inference time. The numerical ratios indicate how average accuracy changes conditional to the decision associated with the node. The symbols below the leaf nodes of the decision tree show the line style used in \autoref{figure:ablationGraph} to allow easily referencing the underlying experiments in detail. 
    }
    \label{figure:ablationTree}
\end{figure*}

We extend the previous ablation study to verify our theoretical analysis of the dual effect of Laplacian dropout. 
\autoref{figure:ablationGraph} introduces a set of experiments that relies on \textit{fair quality} approximate Whittaker-Shanon smoothing kernels (in the middle row of graphs in red), and on \textit{poor quality} truncated Gaussian smoothing kernels (in the bottom row of graphs in blue). \autoref{figure:ablationTree} displays these same results in the form of a decision tree to help recognize the trends that are relevant to our discussion. This decision tree factors the impact of choosing a specific filter quality, choosing whether to use Laplacian dropout or not, and choosing whether to use adaptation or not. This analysis considers accuracy averaged over all resolutions and all datasets as the metric of interest. The numerical values displayed on each node correspond to the average multiplicative change in the metric once a decision is made. We claimed in \autoref{method:laplacianDropout} that Laplacian dropout has two distinct purposes: it regularizes for robustness across a wider distribution of resolutions; it also mitigates numerical discrepancies caused by approximate smoothing kernels when using adaptation. We have demonstrated the first effect in \autoref{experiments:robustness} and can also observe this effect clearly in the decision tree. We can only observe the second effect if we consider the choice of smoothing kernel, which motivates this experimental setup. Our method will conform to our theoretical interpretation if absence of error terms at training time yields better performance in the absence of error terms at inference time, and vice versa; that is to say on the last two levels of the decision tree, on the black and white nodes, tracing a path across two nodes of identical colour should yield a multiplier greater than $1$ at the last node, and conversely, tracing a path across two nodes of opposing colour should yield a multiplier smaller than $1$ at the last node. Our method displays exactly this behaviour, with the discrepancy at the last level of the decision tree growing monotonically with decreases in filter quality. Our method therefore leverages Laplacian dropout not just to improve robustness across a wider distribution of resolutions, but to compensate numerical errors induced by imperfect smoothing kernels, which enables the use of computationally greedy implementations.

\section{Discussion}

We have introduced \methodName{}s, a class of adaptive-resolution architectures that inherits the simplicity of fixed-resolution methods, and the robustness and computational efficiency of adaptive-resolution methods. \methodName{}s substitute standard residuals with \textit{Laplacian residuals} which allow creating adaptive-resolution architectures using only fixed-resolution layers, and which allow skipping computations at lower resolutions without compromise in numerical accuracy. \methodName{}s also implement \textit{Laplacian dropout}, which allows training models that perform robustly at a wide range of resolutions. 

\paragraph{Future Work.} We have provided evidence on classification tasks over low-resolution and medium-resolution image data; our method's ability to generalize is well supported by theoretical justification, but further experiments that include more challenging tasks and high-resolution data are desirable.
We have investigated a form of Laplacian residual that \textit{decreases} resolution, but Laplacian residuals may be generalized to a form that \textit{increases} resolution, which would allow constructing a greater variety of architectures.
We have applied our method in two dimensions on image data, but it is theoretically valid with any number of dimensions; its application to audio data and volumetric data is of interest.

\paragraph{Impact Statement.} This paper presents work whose goal is to advance fundamental research in the field of deep learning and machine learning. No specific real-world application is concerned, although this contribution may render certain forms of technology more accessible.

\bibliography{main}
\bibliographystyle{tmlr}

\newpage
\appendix
\section{Appendix}

\subsection{Background}
\label{background:signals}
We survey fundamental concepts of signal processing and introduce our notation. We aim to provide meaningful intuitions for readers who are not familiar with these principles, and to also rigorously ground our method and satisfy readers who are knowledgeable in this topic. 

\begin{figure*}[h]
    \centering
    \includegraphics[width=0.6\textwidth, page=3]{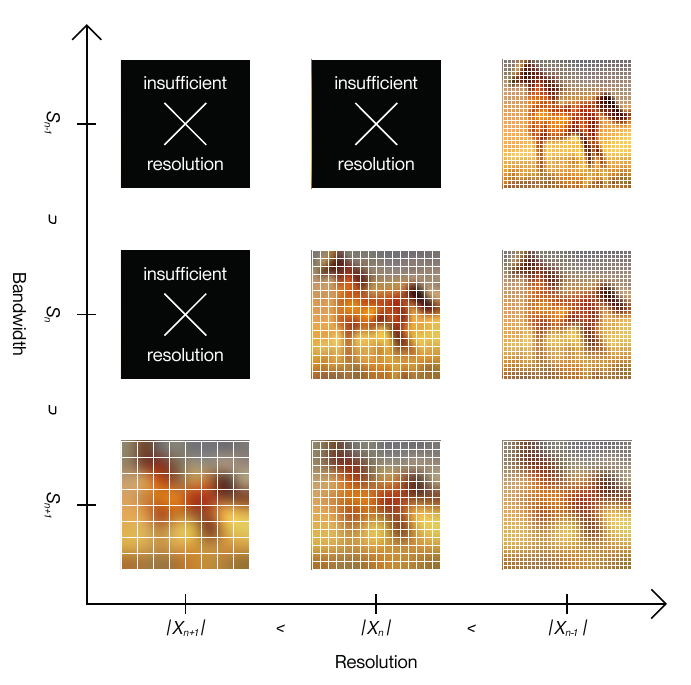}
    \vspace{-1em}
    \caption{
        Diagram showing the same signal captured at three distinct resolutions $|X_{n + 1}| < |X_n| < |X_{n - 1}|$ (laid out horizontally) and three distinct bandwidths $S_{n + 1} \subset S_n \subset S_{n - 1}$ (laid out vertically). Intuitively, resolution corresponds to the coarseness of the grid that is laid over the continuous signal to sample it into a discrete signal, while bandwidth corresponds to the smoothness of the continuous signal. A continuous signal may only be captured into a discrete signal if resolution is sufficiently large to account for bandwidth; the blacked out signals show cases where this condition is not met.
    }
    \label{figure:signalIntuition}
    \vspace{-1em}
\end{figure*}

In \autoref{figure:signalIntuition}, we illustrate how the \textit{resolution} and \textit{bandwidth} of signals intuitively relate to each other while also highlighting the notation and indexing scheme we use to designate these characteristics in our analysis. In the paragraphs that follow, we break down these notions in greater detail.

\paragraph{Signals in continuous form.} We can represent signals as functions $s(\ \cdot \ ) : \mathbf{X} \to \mathbb{R}^f$ that map a \textit{continuous} spatial domain $\mathbf{X}$ that is a nicely behaved subset of $\mathbb{R}^d$ to a feature domain $\mathbb{R}^f$. This representation is useful because it allows us to leverage notions from functional analysis and calculus, and because it is fully independent from the way a signal is captured. We will often refer to the \textit{bandwidth} of signals in this work, which we can intuitively relate to the smoothness of continuous signals; low bandwidth signals are smooth, while high bandwidth signals are detailed.
\paragraph{Signals in discrete form.} We can also represent signals as functions $s[\ \cdot \ ]_n : \mathbf{X}_n \to \mathbb{R}^f$ that map from a \textit{discrete} spatial domain $\mathbf{X}_n \subset \mathbf{X}$ to a feature domain $\mathbb{R}^f$.  This representation enables us to perform computations on signals as they are fully defined by a finite amount of information given by the $|\mathbf{X}_n|$ individual samples $\mathbf{x}_i \in \mathbf{X}_n$. We commonly point to the quantity of samples as the \textit{resolution} of a discrete signal. We associate indexing by $n$ to distinct resolutions throughout this work, where resolution \textit{decreases} as $n$ \textit{increases}, meaning $|X_n| > |X_{n + 1}|$.
\newpage
\paragraph{Casting continuous signals into discrete signals by sampling.} We can easily take a continuous signal $s(\ \cdot \ )$ and create a discrete signal $s[\ \cdot \ ]_n$ by sampling values at points $\mathbf{x}_i \in \mathbf{X}_n$. We notate this process as the \textit{sampling operator} $\mathfrak{S}_n : (\mathbf{X} \to \mathbb{R}^f) \to (\mathbf{X}_n \to \mathbb{R}^f)$:
\begin{equation}
    \mathfrak{S}_n\{s(\ \cdot \ )\} = \mathbf{x}_i \mapsto s(\mathbf{x}_i) \ \ \forall \mathbf{x}_i \in \mathbf{X}_n
\end{equation}
\paragraph{Casting discrete signals into continuous signals by interpolation.} We can reverse the process above and derive a continuous signal $s(\ \cdot \ )$ from a discrete signal $s[\ \cdot \ ]_n$ by applying a convolution with a smoothing kernel $\phi_n$, more formally known as a Whittaker-Shannon kernel \citep{samplingInterpolation, samplingCardinal}. We note that sampling and interpolation are only inverses of each under certain important conditions we come back to later. We interpret the effect of the convolution against a smoothing kernel as filling in the gaps between the samples. We notate the process outlined here as the \textit{interpolation operator} $\mathfrak{I}_n : (\mathbf{X}_n \to \mathbb{R}^f) \to (\mathbf{X} \to \mathbb{R}^f)$:
\begin{equation}
    \mathfrak{I}_n\{s[\ \cdot \ ]_n\} = \mathbf{x} \mapsto \sum_{\mathbf{x}_i \in \mathbf{X}_n} s[\mathbf{x}_i]_n \phi_n(\mathbf{x} - \mathbf{x}_i) \ \ \forall \mathbf{x} \in \mathbf{X}
\end{equation}
\paragraph{Restricting the bandwidth of signals.} We need a slightly more formal way of designating signals that respect certain bandwidth constraints in order to better discuss the equivalence between discrete signals and continuous signals. We can use the smoothing kernels $\phi_n$ we just introduced to define sets of continuous signals $S_n = \{s | s \ast \phi_n = s\}$ that are already smooth enough to be left unchanged by the action of a smoothing kernel. We underline that applying a smoothing kernel $\phi_n$ onto a signal $s$ restricts its bandwidth such that it belongs to the corresponding set of signals $S_n$, which is especially useful as it allows us to guarantee a form of consistency between continuous signals and discrete signals, as we discuss next. We finally note that bandwidth constraints form an ordering $S_n \supset S_{n + 1}$, meaning any signal $s$ that respects a low bandwidth constraint $S_{n + 1}$ also respects an arbitrarily high bandwidth constraint $S_n$.
\paragraph{Equivalence of continuous signals and discrete signals.} We can use discrete signals or continuous signals to designate the same underlying information when the Nyquist-Shannon sampling theorem is satisfied \citep{samplingShannon, samplingMultidimensional}. This theorem intuitively states that a continuous signal with high \textit{bandwidth} requires a discrete signal with correspondingly high \textit{resolution} for sampling to generally be feasible without error. This theorem more formally states that a discrete signal $s[\ \cdot \ ]_n$ can uniquely represent any continuous signal $s(\ \cdot \ )$ that respects the bandwidth constraint encoded by membership to $S_n = \{s' | s' \ast \phi_n = s'\}$, where the expression for the smoothing kernel $\phi_n$ is depends on the sampling density $|\mathbf{X}_n| / \text{vol}(\mathbf{X})$ of the discrete spatial domain over the continuous spatial domain and on the assumptions on the boundary conditions of the continuous spatial domain $\mathbf{X}$ \citep{samplingInterpolation, samplingCardinal, samplingMultidimensional}. We summarize the Nyquist-Shannon sampling theorem by stating that the sampling operator and interpolation operator are only guaranteed to be inverses of each other when the bandwidth constraint is satisfied:
\begin{align}
    \label{equation:discretizationCorrectness}
    s \in S_n \implies \mathfrak{I}_n\{
        \mathfrak{S}_n\{
            s
        \}
    \} = s
\end{align}

\paragraph{Equivalence of operators acting upon continuous signals and discrete signals} We can extend the notion of equivalence between \textit{continuous signals} $s(\ \cdot \ ) : \mathbf{X} \to \mathbb{R}^f$ and \textit{discrete signals} $s[\ \cdot \ ]_n : \mathbf{X}_n \to \mathbb{R}^f$ to encompass the actions that can be performed on the same signals using \textit{operators on continuous signals} $\mathfrak{O} : (\mathbf{X} \to \mathbb{R}^f) \to (\mathbf{X} \to \mathbb{R}^f)$ and \textit{operators on discrete signals} $\mathfrak{O}_n : (\mathbf{X}_n \to \mathbb{R}^f) \to (\mathbf{X}_n \to \mathbb{R}^f)$. We are especially interested in this notion as it enables us to think of our neural architecture as a chain of operators that act on continuous signals which can be cast to act on discrete signals of any specific resolution $|X_n|$. We often see this property formally labeled as \textit{discretization invariance} in the neural operator community and highlight this concept is key to other works which allow adaptation to different resolutions \citep{neuralOperatorFourier, neuralOperator, neuralOperatorSpectral, neuralOperatorDiscretizationInvariance}. We can formally express the equivalence between the continuous form $\mathfrak{O}$ and discrete form $\mathfrak{O}_n$ of some operator as commutativity over the sampling operator when the bandwidth constraint is satisfied:
\begin{equation}
    \label{equation:discretizationInvariance}
    s \in S_n \implies \mathfrak{S}_n\{\mathfrak{O}\{s\}\} = \mathfrak{O}_n\{\mathfrak{S}_n\{s\}\}
\end{equation}
\subsection{Related works}
\label{related:adaptiveDomain}

We cover an additional approach to adaptive-resolution architectures in this complementary section, with the aim of better motivating and situating our approach relative to more mainstream approaches.

\paragraph{Adaptive-resolution by narrowing or widening the domain of evaluation.} 
In principle, many mainstream architectures are able to operate at various resolutions directly, as is the case for fully convolutional architectures. This is because the layers that compose them are invariant or equivariant to translation or permutation, and can therefore evaluate on a wider or narrower domain. In practice, this form of extension to arbitrary resolution is flawed in its ability to guarantee robustness, as it addresses changes in resolution $d|X_n|$ as changes in the bounds of the spatial domain $d\text{vol}(X)$ given fixed spatial sampling density $d(|X_n|/\text{vol}(X)) = 0$.
Operating at a smaller resolution, therefore, means operating on a narrower region of space, which erodes the ability to infer global structure and leads to a collapse in performance when pushed beyond a breaking point. Operating at a larger resolution likewise means operating on a wider region of space, which leads to incoherence once pushed beyond the scope of global structure observed during training. This shortcoming can be sidestepped by interpolating the resolution given at inference to the resolution used at training, which ensures the region of space that is operated on remains consistent. This provides stronger robustness to diverse resolutions, however it amounts to treating these architectures as fixed-resolution and fails to address the problem of computational efficiency.
In contrast, the adaptive-resolution architecture we introduce in our contribution and the other two forms of adaptive-resolution architectures we discuss in \autoref{background} equate changes in the resolution $d|X_n|$ with changes in spatial sampling density $d(|X_n|/\text{vol}(X))$ given fixed bounds of the spatial domain $d\text{vol}(X) = 0$.
Under this more nuanced framing, the region of space that is observed is always constant regardless of resolution, and global structure remains consistently accessible, which guarantees stronger robustness without sacrificing computational efficiency.
\subsection{Experiments}
\label{experiments:details}
We provide further details on our experiments in this section. 

\begin{figure}[h!]
    \centering
    \includegraphics[width=\widthDiagramAnnex, page=4]{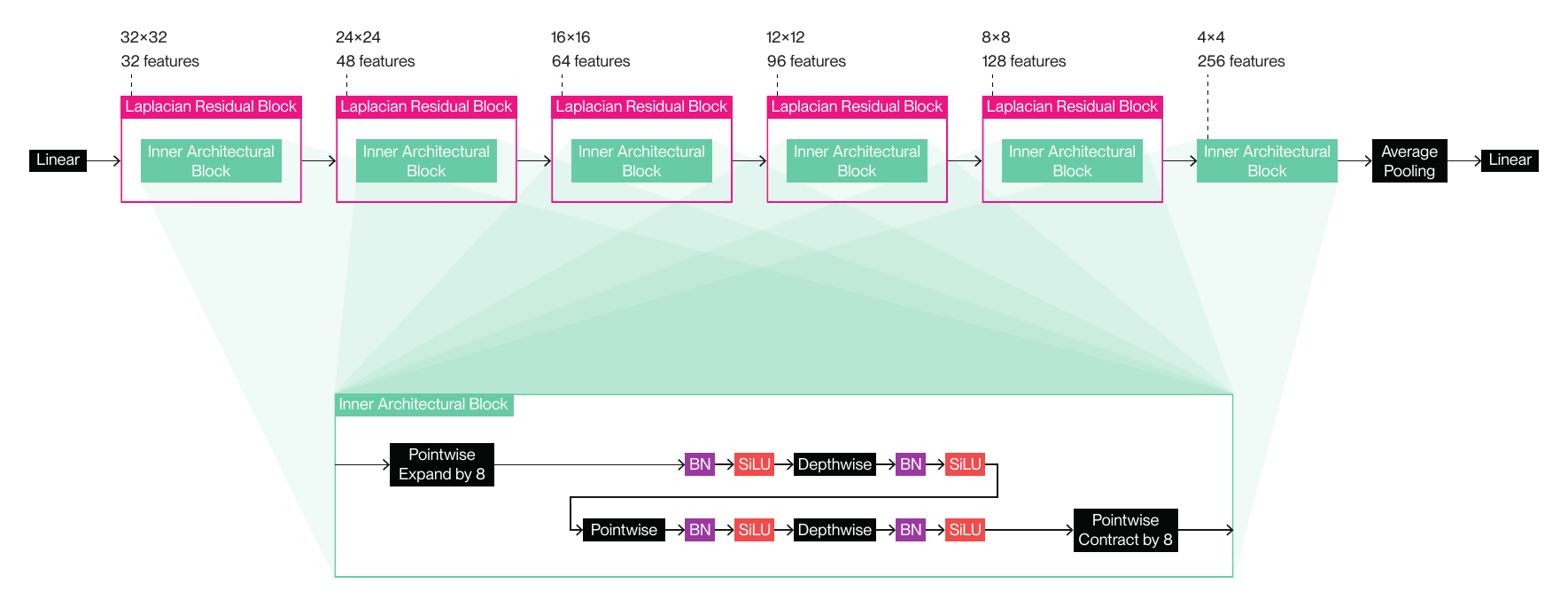}
    \caption{Architecture design for our method on CIFAR10.}
    \label{figure:architectureDesignCIFAR10}
\end{figure}

\begin{figure}[h!]
    \centering
    \includegraphics[width=\widthDiagramAnnex, page=5]{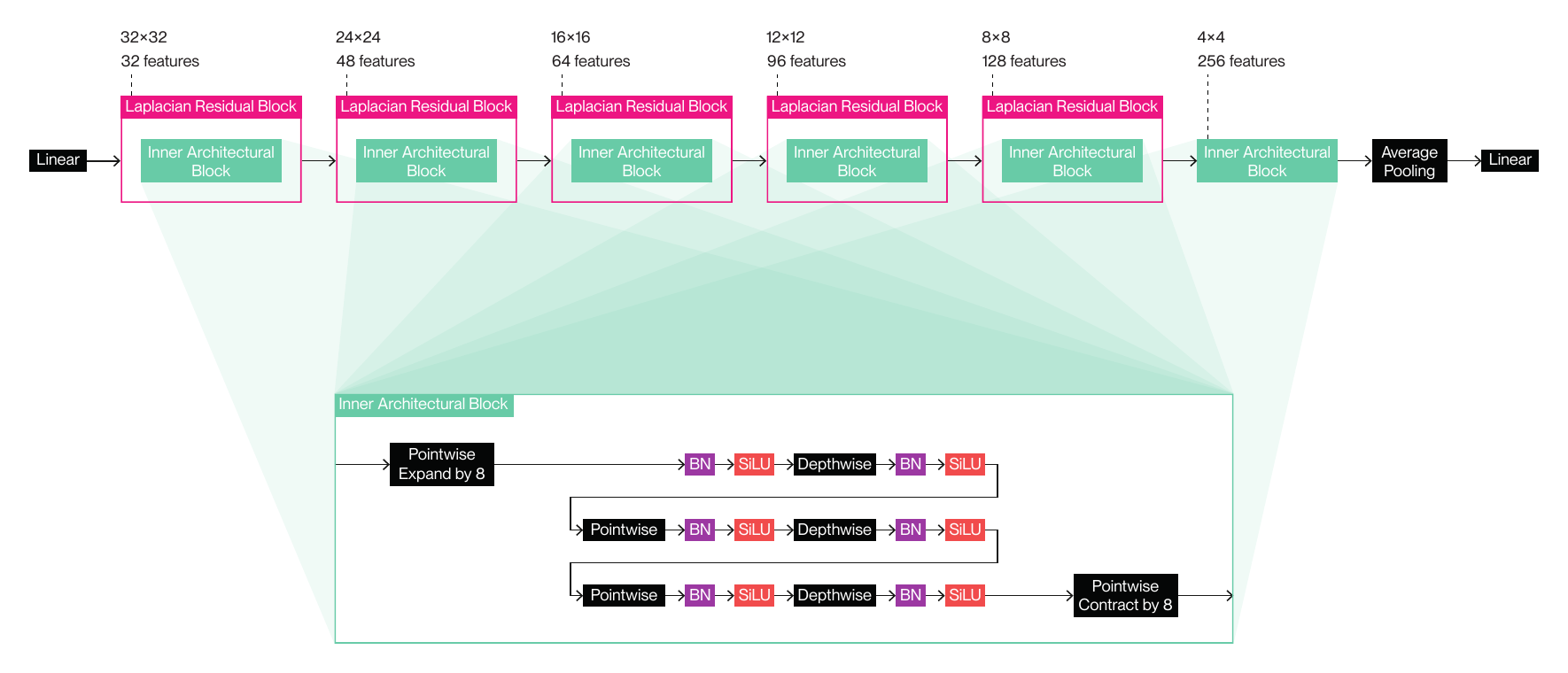}
    \caption{Architecture design for our method on CIFAR100.}
    \label{figure:architectureDesignCIFAR100}
\end{figure}

\begin{figure}[h!]
    \centering
    \includegraphics[width=\widthDiagramAnnex, page=6]{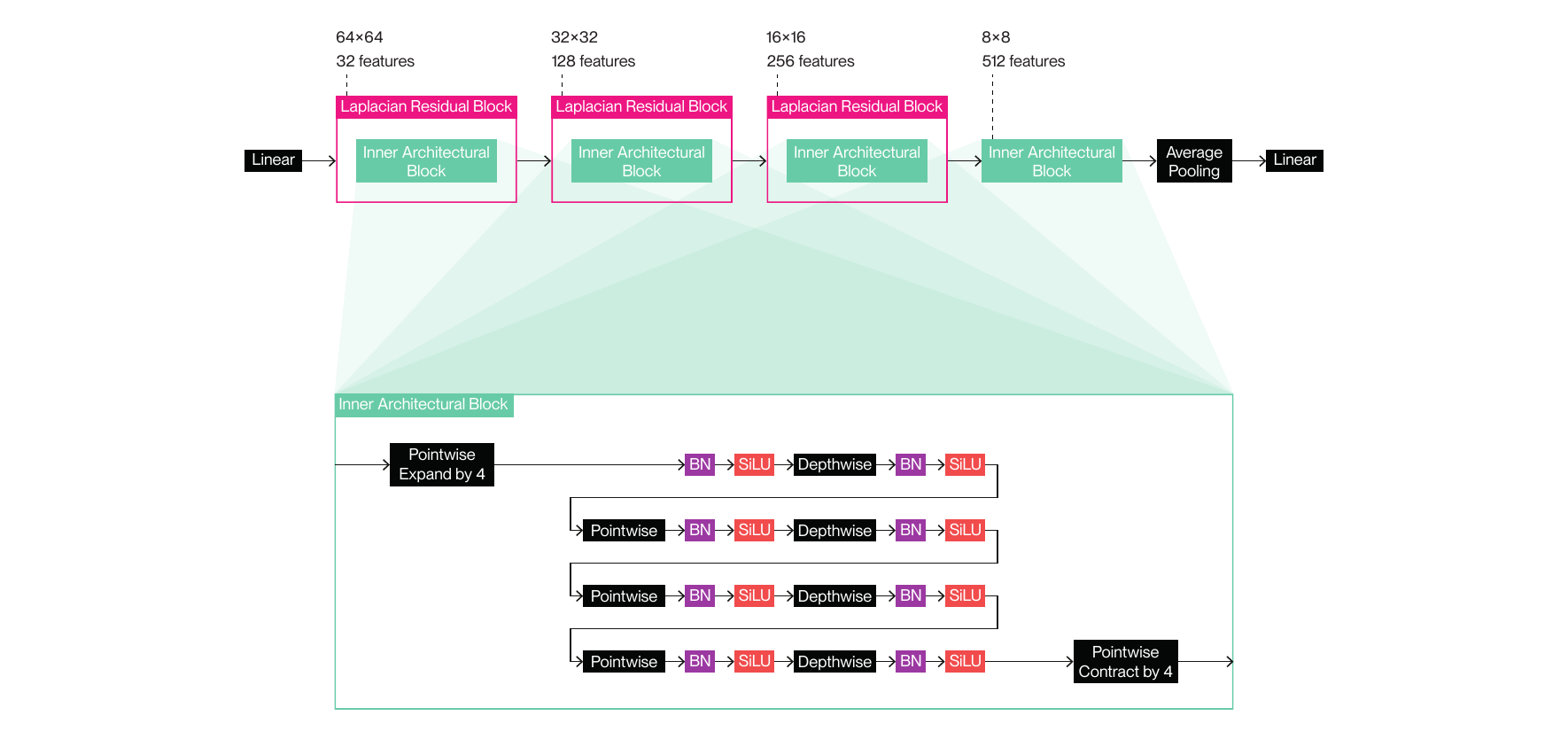}
    \caption{Architecture design for our method on TinyImageNet.}
    \label{figure:architectureDesignTinyImageNet}
\end{figure}

\begin{figure}[h!]
    \centering
    \includegraphics[width=\widthDiagramAnnex, page=7]{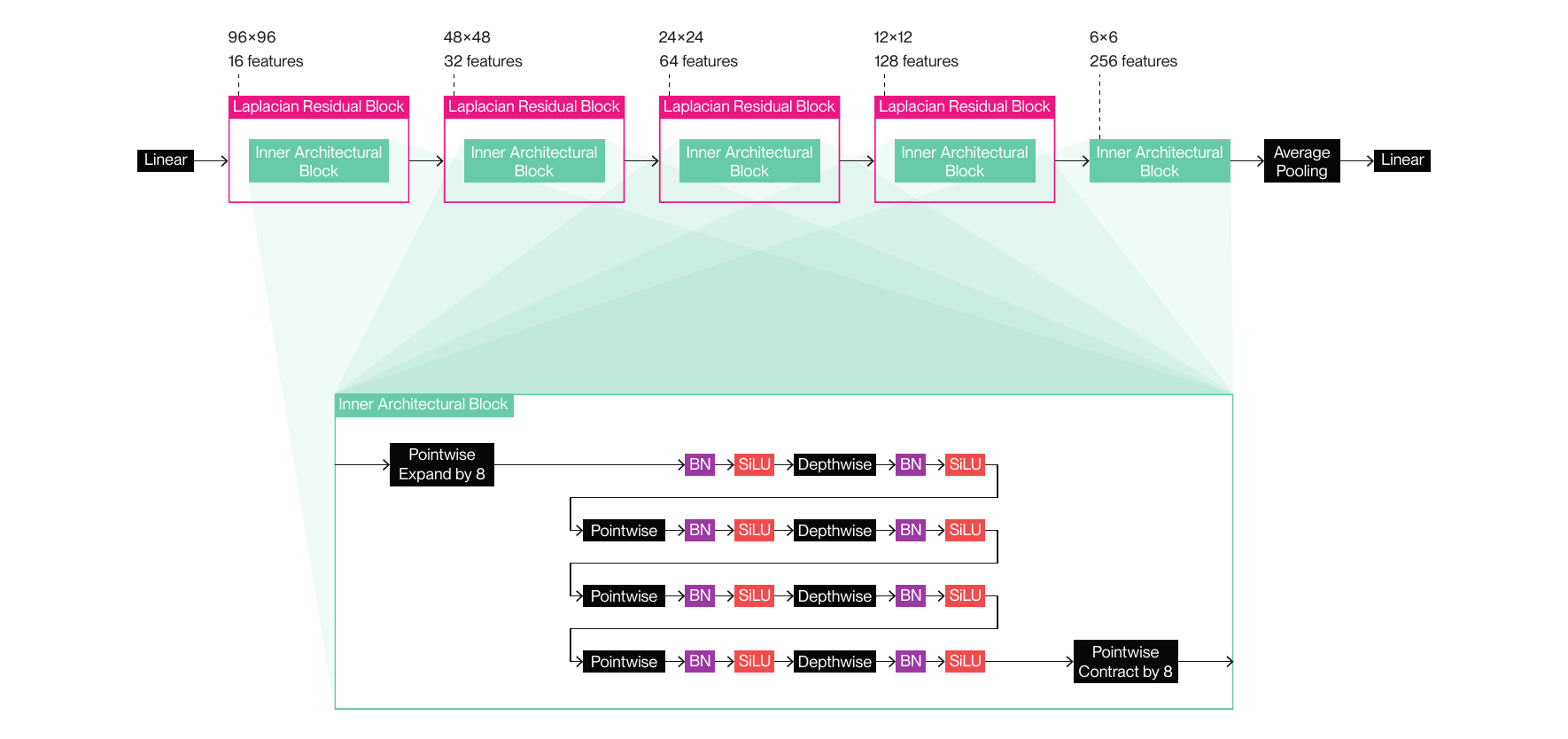}
    \caption{Architecture design for our method on STL10.}
    \label{figure:architectureDesignSTL10}
\end{figure}

\paragraph{Model design.} We provide detailed illustrations for the full architecture designs we use with our method according to each dataset: \autoref{figure:architectureDesignCIFAR10} for CIFAR10 (5.33M-8.09M); \autoref{figure:architectureDesignCIFAR100} for CIFAR100 (9.59M-14.5M); \autoref{figure:architectureDesignTinyImageNet} for TinyImageNet (15.0M-19.8M); and \autoref{figure:architectureDesignSTL10} for STL10 (13.8M-18.4M). We indicate not a single parameter count, but a range of parameter counts for each architecture design, as adaptation enables computation of the forward pass or backward pass using a variable subset of the underlying parameters. We follow a general design pattern inspired by MobileNetV2 \citep{convMobileNetV2} and EfficientNetV2 \citep{convEfficientNetV2} to derive these architecture designs. We nest inner architectural blocks ($b_n$ in \autoref{equation:residualInner}) within a series of Laplacian residual blocks of decreasing resolution and increasing feature count. We create these inner architectural blocks by composing depthwise $3 \times 3$ convolutions and pointwise $1 \times 1$ convolutions in alternation. We set all depthwise convolutions to use edge replication padding to satisfy \autoref{equation:residualInnerConstraint} and ensure resolution remains fixed within each Laplacian residual block. We prepend this string of layers with a pointwise convolution that expands the feature channel count. We conversely terminate the sequence of layers with a pointwise convolution that contracts the feature channel count inversely. We separate each convolution with a batch normalization \citep{batchNormalization} and a SiLU activation function \citep{activationSiLU}, chosen for its tendency to produce fewer aliasing artifacts. We apply different Laplacian dropout rates ($p_n$ in \autoref{equation:residualDropoutIndep}) depending on the dataset: $0.6$ for CIFAR10; $0.3$ for CIFAR100; $0.3$ for TinyImageNet; $0.3$ for STL10. We use a common classification head that consists of a single linear layer with a dropout set to $0.2$, which is applied after global average pooling. The designs were chosen by sweeping over different configurations for inner architectural blocks, and over different resolutions and number of features for the Laplacian residual blocks that contain them. The number of permutations per final sweep ranged between 18 to 216 for each dataset. 

\newpage
\paragraph{Model training hyperparameters.}
We provide the specific hyperparameters used during training. 

For CIFAR10 and CIFAR100, across all methods, we use AdamW \citep{optimizerAdamWeightDecoupled} with a learning rate of $10^{-3}$ and $(\beta_1, \beta_2) = (0.9, 0.999)$, cosine annealing \citep{optimizerWarmRestarts} to a minimum learning rate of $10^{-5}$ in 100 epochs, weight decay of $10^{-3}$, and a batch size of $128$. We use a basic data augmentation consisting of normalization, random horizontal flipping with $p = 0.5$, and randomized cropping that applies zero-padding by 4 along each edge to raise the resolution, then crops back to the original resolution. 

For TinyImageNet and STL10, across all methods, we use SGD with a learning rate of $10^{-2}$, cosine annealing \citep{optimizerWarmRestarts} to a minimum learning rate of $0$ in 100 epochs, weight decay of $10^{-3}$, and a batch size of $128$. We use TrivialAugmentWide \citep{augmentTrivialWide} to augment training. 

\end{document}